# AI Enabled Maneuver Identification via the Maneuver ID Challenge

| Kaira Samuel | Matthew LaRosa | Kyle McAlpin | Morgan Schaefer | Brandon Swenson |
|---|---|---|---|---|
| MIT | USAFA | USAF | MIT | USAF |
| Cambridge, MA | Colorado Springs, CO | Cambridge, MA | Cambridge, MA | Ft. Meade, MD |
| kmsamuel@mit.edu | c23matthew.larosa@usafa.edu | kmcalpin@mit.edu | meschae@mit.edu | bmswens@mit.edu |

| | Devin Wasilefsky | Yan Wu | Dan Zhao | Jeremy Kepner |
|---|---|---|---|---|
| | USAFA | MIT | NYU | MIT |
| | Colorado Springs, CO | Cambridge, MA | New York, NY | Cambridge, MA |
| | c23devin.wasilefsky@afacademy.af.edu | yanswu@mit.edu | dz1158@nyu.edu | kepner@ll.mit.edu |

## INTRODUCTION

Flight simulators play a critical role in pilot training. Current training paradigms require scarce, highly-experienced instructor pilots to teach even the most basic flight maneuvers, beginning with basic flight maneuver familiarization in flight simulators. AI has significant potential to enhance simulator-based training by providing real-time feedback on the quality of each flight maneuver to student pilots for early-stage learning. An important first step towards achieving AI enhanced pilot training is teaching an AI to recognize categories of flight maneuvers from flight simulator data. The application of AI to any new domain is a daunting task. One proven approach is to develop an AI challenge to grow the AI ecosystem around a new domain.

AI challenges serve as a primary tool for development and innovation by engaging the broader research community. Challenges have been created to advance the fields of machine learning, high performance computing, and visual analytics. Introducing important AI problems to the public allows for increased collaboration among diverse research teams and maximization of potential solutions (see Figure 1). Challenges such as YOHO (J.P. Campbell, 1995), MNIST (C.C.Y. LeCun and C.J. Burges, 2017), HPC Challenge (*HPC Challenge,* 2017), Graph Challenge (E.Kao et al., 2017; J. Kepner et al., 2019; S. Samsi et al., 2017), ImageNet (O. Russakovsky et al., 2015) and VAST (K.A. Cook et al., 2014; J Scholtz et al., 2012) have driven major developments in various fields. Each of these challenges have catalyzed critical research efforts in their respective fields: YOHO enabled speech research; MNIST remains foundational to the computer vision research community after two decades; HPC Challenge has stimulated research on parallel programming environments and plays a critical role in supercomputing acquisitions; Graph Challenge has produced award-winning software and hardware; ImageNet has enabled vision research; VAST challenges the visual analytics research community with new topics annually. These and many other challenges impact their fields immensely by introducing bedrock tools for acquisition and source selection processes, as well as providing a baseline for future challenges in the field.

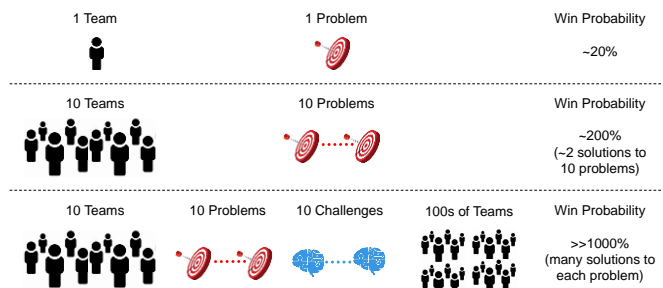

**Figure 1: AI Challenge Motivation**. AI challenges allow diverse teams to work on common well-defined problems and greatly increase the probability of finding a solution.

Interdisciplinary research teams at the Department of the Air Force-Massachusetts Institute of Technology AI Accelerator (DAF-MIT AI Accelerator) composed of Air Force and Space Force personnel, MIT researchers, and MIT Lincoln Laboratory technical staff, create public challenges to build AI ecosystems around difficult Air and Space Force problems, which are discussed in more detail at aia.mit.edu/challenges. These challenges seek to make fundamental advances in AI to support broader societal needs like AI explainability and robustness. We introduce one of these AI Accelerator challenges, the Maneuver Identification (ID) Challenge, targeting increased efficiency in pilot training pipelines.

**Data Provenance**

The dataset provided in the challenge represents recordings of student training flights conducted in virtual reality (VR). At the time of collection, PTN conducted US Air Force Undergraduate Pilot Training at Randolph Air Force Base, TX, primarily in the T-6A Texan II training aircraft. As an experimental pilot training unit, PTN's early victories included the incorporation of virtual reality training hardware and methods into the flying training syllabus. Due to the data collection occurring in live, dynamic flying training operations, some meta-data associated with the dataset was unavailable. Some unknown meta-data includes the identity and experience level of the pilot flying in the recording (brand new student or highly experienced instructor pilot), the number and types of maneuvers contained in each recording file if any, the quality of each maneuver contained in each recording (unsatisfactory, fair, good, or excellent), the formality of the training conducted in the recording (formal syllabus training event where students are graded on performing maneuvers in accordance with syllabus standards, informal unguided practice- "play", or instructor pilots practicing maneuvers), the start and stop times or location of any maneuver(s) contained in each recording, the type and number of discontinuities in flight path (intentional discontinuities such as intentional repositioning of the simulated aircraft to accomplish training objectives, unintentional discontinuities such as a simulator artifact, or no discontinuities). Additionally, the data needed to be cleaned and curated to be usable and releasable. Cleaning actions performed included the exploration of the dataset, removal of duplicate records, identification and verification of the type of data contained in each field or column, removal of unnecessary or redundant columns, identification of the units of measure for each column, conversion of data types to internationally recognized units of measure (i.e., converting nautical miles per hour to meters per second), and moving each recording to a standard starting location and altitude (i.e., removing absolute simulated location and altitude of the aircraft).

**Previous Relevant Research**

The Air Force began Pilot Training Next (PTN) to explore the use of new technologies, such as consumer virtual reality flight simulators and novel instructional tools, with small test cohorts to personalize training and improve student access to training resources (J. Stockton, 2019). The Air Force Chief Data Officer (SAF/CO) worked with PTN researchers to begin the Maneuver ID effort; early maneuver detection attempts involve comparing sections of flight data to known exemplary maneuvers to detect the maneuvers being attempted (J. Stockton, 2019). The data used in SAF/CO efforts is being repurposed for the Maneuver ID Challenge to continue the influx of novel solutions. The body of work surrounding maneuver identification and flight pattern learning using artificial intelligence is significant and a few examples are noteworthy. In the 1970s, NASA developed a technology implementing Adaptive Maneuver Logic (AML) to virtually simulate air-to-air combat. AML tracks the flight patterns of its opponents to predict what their next move will be using bank angle, lift, and thrust (G.H. Burgin, 1975). This has paved the way for more maneuver identification research: a 2015 project, motivated by the importance of load analyses on aircraft, analyzed flight data using a library of maneuvers (Y. Wang et. al., 2015). Previous work has explored the application of AI in classifying and predicting driving maneuvers like lane changes using position and velocity data (A. Benterki et al., 2020; Y. Wang and I. W. Ho, 2018) as well as the classification of unsafe driving maneuvers from dashcam videos (M. Simoncini et al., 2022). Using Bayesian methods, random trees, and various neural networks, the technology can predict lane changes in cars with up to 97% accuracy given six-second sequences of data input (A. Benterki, 2020).

**Team Contributions**

The development of any AI challenge in a novel domain is an ongoing, iterative process. As teams attempt the variety of AI tasks available, issues are detected and addressed to improve the challenge. The feedback and solutions from AI researchers received in the initial stages of these challenges are a crucial aspect of developing an AI challenge. This paper presents the contributions of four initial research teams that have taken different approaches to the Maneuver

ID Challenge. The teams are making their software available to the community for others to build upon and have discovered several areas of improvement that are discussed within this manuscript. As is typical in such AI challenges, the variety of solutions produced provide a foundation for future solutions.

The major contributions of these teams are described in the remainder of the paper and are summarized as follows:
- Initial observations of the data yielded additional data labels identifying various features within the sorted data that contribute to the growing body of information on the data.
- Effective sorting of "good" and "bad" data which is essential for scaling up the dataset has been achieved using several AI techniques, including simple statistical measures, random forest models, bagging trees, simple decision trees, support vector machines, logistic regression, and convolutional neural networks.
- Maneuver identification has also been approached using the aforementioned techniques.
- Various unsupervised learning models have also been applied to circumvent the low-shot nature of truth-labeled data. This low-shot labeling challenge has also led to one of our teams to develop new web-based labeling services using video simulations created from the trajectory data. Such services can enlist the expertise of pilots and those familiar with Air Force maneuvers to acquire more labeled data once it is ready to be released.

**CHALLENGE DESCRIPTION**

PTN gathered thousands of distinct pilot training sessions from hundreds of hours on flight simulators, flown by students, trained pilots, and instructors. The sessions were flown using the Lockheed Martin Prepar3d flight simulator software (prepar3d.com) (J. Stockton, 2019). Commercial flight simulators of this type are not designed to dynamically emit data in a multi-pilot training context. PTN developed a novel state-of-the-art data logging system that can aggregate data from multiple simulators simultaneously. Following data transformations and anonymization making the data AI ready, the trajectory data has been made available in a Dropbox folder as tab separated value (TSV) files. The process of data transformation is detailed in this Challenge's previous paper, *Maneuver Identification Challenge,* found at Maneuver-ID.mit.edu/Motivation. The TSV files consist of a plain text table containing the times, positions, velocities, and orientations of the aircraft throughout the flight session (i.e., see Figure 2). The files can be read by most data processing systems and viewed in any spreadsheet program. In addition, 2-dimensional top-down views of ground-track trajectories are stored as portable network graphics (PNG) files (e.g., see Figure 3).

**Figure 2: Example TSV file for a single flight simulator recording**

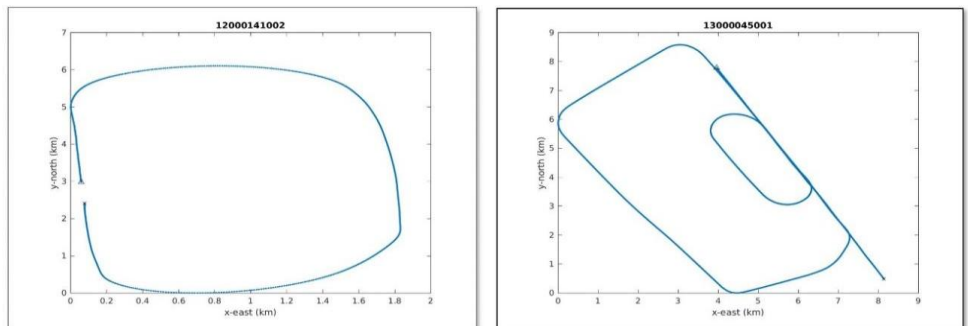

**Figure 3: Example PNG files for flight simulator recordings**

The Maneuver ID dataset consists of several different artifacts that provide information and resources for those interested in participating in the challenge. The dataset containing the unlabeled flight simulator recordings along with is hosted on DropBox- it can be accessed by navigating to Maneuver-ID.mit.edu/data and signing the data sharing

I/ITSEC 2022 Paper No. 22475 Page 5 of 12

agreement located there. The agreement follows standard data-sharing best practices, notifying participants of the anonymization of the data, confirming the legitimacy of their research, as well as ensuring that they agree to respect the anonymization of the data. The webpage at Maneuver-ID.mit.edu contains other supporting artifacts including a challenge description and background, a list of 18 maneuvers with maneuver parameters and narrative How-To descriptions, and videos of trained pilots executing some of the described maneuvers in accordance with AETCMAN 11-248, T-6 Primary Flying (HQ AETC/A3VU, 2016).

The Maneuver ID Challenge has identified three tasks for the community.
1. Sorting the physically feasible (good) and physically infeasible (bad) data into separate sets based on the presence of unbroken trajectories (good) and identifiable maneuvers (good) and straight lines (bad), jumps (bad), and maneuvers that break physical laws (bad). The good data consists of sorties that each have a collection of realistic maneuvers flown, whereas the bad data contains sorties with inconsistencies that do not align with feasible, real world flight patterns. Currently, there is good and bad trajectory truth data that has been sorted and verified manually by two non-subject matter experts (non-SMEs) and spot checked by SMEs (i.e., pilots), meaning errors in these labels may exist. The manually sorted truth data, which separates the good and bad files into different folders, is included as part of the data set.
2. Identifying which maneuver(s) the pilot is attempting to execute. Each sortie has trajectory data containing several maneuvers; labeled sorties are not currently available, but teams are working on labelling.
3. Scoring the pilot once the maneuver has been identified. This could greatly benefit the efficiency and quality of the pilot training education.

The current body of work from participants focuses on the first two tasks. To address the lack of labeled sorties currently available for the second task, teams have taken different strategies – developing models based on the prospect of future labeled sorties and experimenting with unsupervised models, both of which are discussed later.

## DATA EXPLORATION & PRE-PROCESSING

Gaining intuition on datasets before diving into machine learning model development can pay dividends. We created methods to visualize the flight maneuvers contained in the data and perform additional labelling to gain intuition about this rich, complex dataset. We provide these methods to challenge participants and detail them below.

**Visualization**

One approach to visualizing the flight simulator data involves creating video animations from the raw data (e.g., see Figure 4). Blender's open-source Python library, Blender Python, allows a programmatic approach to creating these animations. After loading an aircraft model (in this case, a T-6 Texan, which is an older version of the T-6A Texan II used in our pilot training simulator data, but is a free, widely-available model), a specified data file is iteratively animated by setting the position and rotation of the airplane. Additional smaller improvements can be adjusted within the Blender environment (e.g., the background and clipping length) to create smoother animations. The animated object can then be rendered or exported as a GL Transmission Format (glTF) or AVI video file format to be viewed outside of the Blender environment.

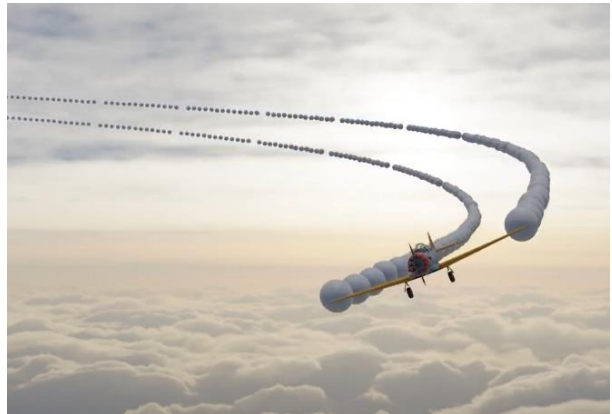

**Figure 4: Screenshot of Blender Animation**

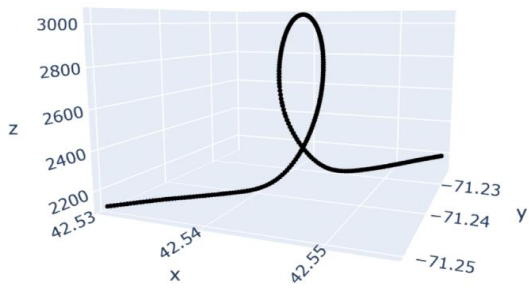

Another visualization approach enlists Python's open-source graphing library, Plotly, to create interactive visualizations. Both 2D and 3D visualizations of the data can be created using Python's Plotly, Chart Studio, and Kaleido. Taking a list of dictionaries containing the data as the input, the data can be plotted in interactive graphs. The 2D visualization offers an aerial visualization of the plotted $x$ (East) and $y$ (North) coordinates, whereas the 3D visualization plots all three position coordinates ($x$, $y$, and $z$) in a 3D coordinate plane (e.g., see Figure 5).

**Additional Labeling**

This new visual analysis allowed aspects of the data that pose challenges to the sorting and categorization of maneuvers to be identified through an automated process. These include the following: impossible speeds, teleportation (i.e., instant repositioning of the plane in the simulator space), long taxiing maneuvers (i.e. aircraft maneuvering on the ground before or after flight), and irregular stops throughout the duration of some 'good' labeled sorties. These irregularities were identified and documented using R. Each file was iterated through to determine if the plane was exhibiting one of the behaviors described based on threshold values. If the plane exhibited a behavior for a certain amount of time, then it was labeled as such. Specifically, stopped and taxiing sorties were found by examining the speeds of the aircraft and determining whether the aircraft was moving. Teleportation (aka "discontinuities") was identified based on discontinuous jumps in the plane's position. Lastly, impossible speeds were identified by differentiating the position, yielding the coordinate speeds, and determining any infeasible values. Based on this data, a spreadsheet was compiled listing each of the files containing irregularities along with the label next to it (i.e., see Figure 6).

Figure 6: First 15 Files Exhibiting Irregularities with Behavior Labels

### TASK 1: SORTING

Once the data is sufficiently curated, visualized, and explored, model development can begin. A variety of approaches exist, including basic statistical measures, canonical data models, and convolutional neural networks. The first task focuses on the challenge of sorting the available data into 'good' and 'bad' categories based on the physical feasibility of a pilot executing the maneuver(s) within the recording. Unlike real aircraft, VR simulators used to collect the data can be left idling continuously to avoid lengthy computer boot-up and loading times. If the simulation was left running while unused, it may have recorded long stretches of no or minimal aircraft movement. Likewise, simulators are often used to re-accomplish a single maneuver many times over, or quickly snap between simulated locations and altitudes to accomplish subsequent training objectives. These simulator behaviors may manifest in the data as 'discontinuities' where the simulated aircraft is instantaneously re-positioned to a new location or state.

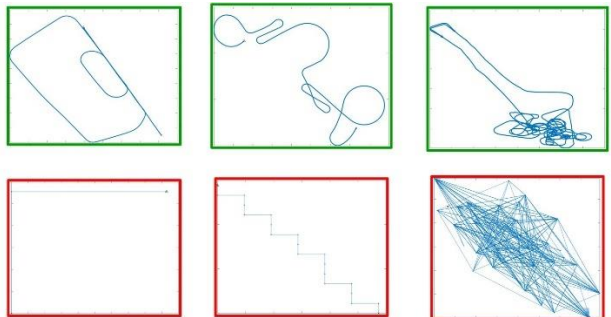

Figure 7: Sample PNG Truth Data Manually Sorted into 'Good' (green) and 'Bad' (red)

Researchers can use the manually sorted 'good' and 'bad' truth data to train and validate their models (i.e., see Figure 7).

**Approach 1: Basic Statistical Measures:** One approach leverages basic statistical measures of mean and standard deviation to score the trajectories. Although simple in its implementation, this approach has performed very well when attempting the first task of sorting the 'good' and 'bad' files. Upper bound values for the mean and standard deviations of the position coordinates and roll of the aircraft were chosen to separate the 'bad' files from the 'good' ones. The algorithm scores each maneuver based on a Boolean operation that determines if each file within the sorted 'good' and 'bad' folders meets the criteria of the statistical measures. The true positive percentages are calculated by dividing the 'good score' (the total number of files that meet the criteria) by the total number of files in the manually labeled good data folder; true negative percentages are calculated by dividing the 'bad score' (the total number of files that do not meet the criteria) by the total number of files in the manually labeled bad data folder. Table 1 contains the results for the tested statistical parameters.

**Table 1: True Positive and True Negative Percentages for Starter Statistical Parameters**

| Mean (m) | Standard Deviation (m) | True Positive (%) | True Negative (%) |
|---|---|---|---|
| xEast < 500 | xEast < 100 | 97.8 | 91.5 |
| yNorth < 500 | yNorth < 100 | | |
| Roll < 0 | | | |

Future work on this approach could optimize the mean and standard deviation limits to achieve higher true positive and negative rates or introduce additional statistical measures beyond mean and standard deviation.

**Approach 2: Canonical Models:** Another sorting approach involves training more sophisticated classifiers. Due to the structural differences between the good and bad sortie data, summary statistics can be aggregated and used to train a multitude of statistical models - the performance results for each classifier are summarized in Table 2. The results are derived from a balanced data set where the training data consists of equal components of feasible and infeasible data. Random Forest and bagging tree models trained on the summary statistics dataset are the best-performing. These two models are able to accurately predict whether a sortie is good or bad with an approximately 98% accuracy, F1 score, specificity, and recall, using the default hyperparameter settings in R. These models do not contain the supplemental data acquired by filtering the 'good' sorties.

**Table 2: Classifier Performance on each Respective Balanced-Test Dataset**

| Dataset | Classifier | Accuracy (%) | Average F1 Score (%) | Average Specificity (%) | Average Recall (%) |
|---|---|---|---|---|---|
| **Summary Statistics** | Random Forest | 98.67 | 98.67 | 98.67 | 98.67 |
| | Bagging Trees | 98.00 | 97.99 | 98.67 | 97.33 |
| | Decision Tree | 94.67 | 94.81 | 92.00 | 97.33 |
| | Support Vector Machine | 92.67 | 93.08 | 86.67 | 98.67 |
| | Logistic Regression | 92.00 | 91.67 | 96.00 | 88.00 |
| | Naïve Bayes | 91.33 | 91.16 | 93.33 | 89.33 |
| | Neural Network | 75.33 | 68.38 | 97.33 | 53.33 |

**Approach 3: Image Classification via Neural Networks:** A final sorting approach is to use image classification. This approach assumes that the 2D top-down visual representation of the sortie's flight path in Figure 7 is sufficient to distinguish between good and bad instances, which in general can be an oversimplification when considering 3-dimensional flight, but nevertheless produces promising results.

A convolutional neural network (CNN) can be constructed and trained to classify the files based on image representations of the labeled data with a high degree of accuracy. Preprocessing may include combining the x and y velocity parameters into one value of airspeed to reduce the number of parameters. A CNN can then be constructed based on the graphs of the 'yNorth' positions vs. the 'xEast' positions of the virtual plane. The labeled data can be split into training and testing categories, with a portion of the training files set aside for validation. A basic CNN can be written assisted by torch.nn. The model can be trained on the PNG files for 5 epochs with a learning rate of 0.01. Validating the model with the testing data results in an overall accuracy of ~95.3%. Future work for this approach may come from changing the size of and applying transformations (i.e., rotation and cropping) to the images, which may reduce speed, but increase accuracy. One can also create a deeper CNN, adding altitude over time as another layer, or borrow a more sophisticated CNN, such as the ResNet50 architecture.

As an image classification task, there exists a wealth of deep vision model architectures that have achieved state-of-the-art performance on diverse image datasets. For this task, we can borrow a ResNet50 architecture and train the model from scratch, without using its original pre-trained weights on ImageNet, on the PNG image files of the sorties' flight path for 10 epochs using stochastic gradient descent with an initial learning rate of 0.01, momentum set at 0.9, weight decay set at 0.00005, and a batch size of 32 images per iteration. Since the dimension of the final dense layer of ResNet50 is greater than the number of classes for this problem, the final dense layer is modified to specifically match its dimension to the number of classes in this problem. The images undergo some standard pre-processing, such as cropping and resizing, and are split into a train-test set of 80-20 where the training images are shuffled every epoch during training to prevent the network from learning incidental information or spurious correlations gleaned through the ordering of the images. Under these settings, the network achieves a test performance of ~98% with a true positive rate of ~99% and true negative rate of ~93% on the test set where the positives correspond to the "good" class and the negatives correspond to the "bad" class based on the previously sorted 'truth' data (see Table 2b). Given the class imbalance (ratio of about 88:12 good to bad files) present in the data, future work should keep in mind the nuances around model evaluation or data supplementation efforts of the under-represented class. However, it is worth noting that the ResNet50, even when trained and tested on the imbalanced dataset, delivers considerable performance.

**Table 2b: ResNet50 Performance on PNG Dataset of Flight Maneuvers, Unbalanced Test Dataset**

| Dataset | Classifier | Accuracy (%) | True Positive Rate (%) | True Negative Rate (%) |
|---|---|---|---|---|
| Images of Maneuvers | ResNet50 (Modified) | 98.6 | 93.3 | 98.8 |

A direction for future work would be to develop a means to distinguish between good and bad sorties in a multimodal fashion: namely, a model that is able to take different representations of sortie data and use them to learn different aspects of good and bad in sortie data across various representation domains. One example is a multi-branch neural network that takes in multiple inputs (i.e., one branch takes in tabular data as input and another takes in images) but, via some form of weight sharing and concatenation in its hidden layers, still performs binary classification in the final layer. This way, the network may be able to glean additional information across different sortie representations and modalities that each representation alone may not sufficiently provide.

**TASK 2: IDENTIFYING MANEUVERS**

The second task involves identifying the maneuvers within the flight recordings that the pilot is attempting to execute. Currently, only single examples of each maneuver, containing a TSV and PNG file for a single maneuver flown on the simulator, are available to train and test with, making many of the traditional approaches to this problem challenging. Furthermore, the sorties may each contain several maneuvers within a single flight, creating a potential additional task of splitting up individual files based on the beginnings and ends of different maneuvers. To mitigate this, one team is building a Mechanical Turk-style labeling application crowd-source labels from experts. This consists of a Django web application with an interactive interface to view and label the glTF files created by Blender. Development of the system is still underway. However, preliminary results field an interactive environment with object tracking and handling animation playback. Further development will create glTF animation markers and store the labeled information in a database.

**Approach 1: Time-series Random Forest:** Approaches relying on the prospect of more labeled data have been set up. A time-series Random Forest approach can be conducted to identify the type of maneuver being flown in a sortie. Python's open-source package sktime can be primarily used to train the model. Following data resampling and cleaning of NaN values, the multivariate model is trained by concatenating the variables. The model is then able to predict the sortie's maneuver; however, the accuracy of the labels suffers from the lack of truth data.

**Approach 2: Unsupervised Learning:** Under the presumption that no labels will be available, it is necessary to explore and develop an approach that does not require labels to generate predictions. Though the quality of these predictions may be found wanting if labels are made available, the goal here is to simply provide a potential framework that can generate predictions, without any supervision, based solely on the similarity between identified maneuvers and maneuvers in sortie data.

To identify maneuvers in sortie data, one method identifies patterns in the exemplar maneuver data and matches them to similar patterns in the challenge data. These similarities can exist on both the univariate and multivariate scale. In terms of univariate similarity, each feature in the maneuver data should exhibit a strong degree of similarity with the same corresponding feature in some part of the sortie data if said maneuver is present. For multivariate similarity, the correlations between the features of a maneuver should also strongly resemble the correlations between the features in some part of the sortie data as well if said maneuver is present. Together, these two similarities can be used to quantify the likelihood that any maneuver is present in the sortie data.

For univariate similarity, dynamic time warping is used to compare maneuver features with their respective counterparts in the sortie data since sorties can contain multiple maneuvers and be of different lengths (usually longer than the maneuver data). Dynamic time warping is used to compare each pair of features as well as their first differences (i.e. the feature minus itself lagged by one time step) to help quantity both raw similarity but also similarity in their differenced forms to help mitigate potential issues with non-stationarity when comparing time series. For multivariate similarity, a correlation matrix distance is used and takes the form of one minus the cosine similarity between two correlation matrices. Together, the univariate and multivariate similarity measures produce a set of similarity scores for each maneuver and its likely presence in each sortie dataset. These scores are transformed via softmax to produce probability scores that are then combined together to identify the likelihood a maneuver is present in a sortie dataset. Ideally, univariate and multivariate similarity measures should be calculated via a rolling window on the sortie dataset for each maneuver since the length of the sortie dataset is typically longer than each maneuver's recorded data and the sortie dataset can contain more than multiple maneuvers; however, due to potential computational burdens, we compute similarities on the full set without a rolling window and leave this approach as a direction for future work.

**REFERENCE CODE DISTRIBUTION**

Each team has generously agreed to contribute their starter code for each of the approaches discussed in this paper. The repository is publicly available in GitHub under the Maneuver-Identification organization. Participants can visit [github.com/Maneuver-Identification](github.com/Maneuver-Identification) to access the starter code and [Maneuver-ID.mit.edu/data](Maneuver-ID.mit.edu/data) to submit the data sharing agreement and gain access to the data via Dropbox.

**SUMMARY AND NEXT STEPS**

AI can significantly improve Air Force pilot training by reducing the need for instructor pilots in the introductory phases of pilot training and giving student pilots access to better tools for self-paced learning in VR. The Maneuver ID Challenge is championing the creation of such capabilities to increase pilot training efficiency by providing data, tools, and starter algorithms to the global public. The Maneuver ID challenge has assembled thousands of virtual reality simulator flight recordings collected by actual Air Force student pilots at Pilot Training Next (PTN). This publicly released novel USAF flight training dataset can be accessed through Maneuver-ID.mit.edu, along with other resources detailing the challenge. This paper has discussed the various AI methods applied in separating "good" from "bad" simulator data as well as categorizing and characterizing maneuvers. The algorithms and software are being released as baseline performance examples for future participants to work off. Table 3 outlines the progression and development of artifacts that have contributed to the set up and continuation of the challenge.

AI challenges iteratively improve throughout their lifetime. The Maneuver ID Challenge will continue to produce artifacts to increase the value and accessibility of the dataset for participating researchers. The link to the GitHub repository containing commented code and explained approaches is available on the website and will serve as a template for ongoing solutions that address the drawbacks of some of these approaches: the appearance of several maneuvers within a single file, the lack of several truth data examples of each maneuver, etc.

**Table 3: Progression of artifacts developed during the Maneuver ID Challenge**

| Original | New | In Development |
|---|---|---|
| Webpage containing pertinent information on the challenge (Maneuver-ID.mit.edu) | 'News' page and 'FAQs' added to website to document major updates to the datasets and questions about the challenge (Maneuver-ID.mit.edu/News, Maneuver-ID.mit.edu/FAQs) | Blender produced gITF files containing animated visual representations of the raw data |
| Dropbox folder containing flight simulator data as well as sample maneuver data | Created GitHub repository for starter code (github.com/Maneuver-Identification) | Django web application that will allow crowdsourced labeling efforts to acquire truth data for Task 2 |
| IEEE Paper published August 2021 outlining the challenge: *Maneuver Identification Challenge* | Spreadsheet documenting additional labels for sorties containing irregularities | |
| YouTube playlist containing video recordings of exemplar maneuvers in the VR flight simulators (Maneuver-ID.mit.edu/maneuvers) | | |

Two outcomes of this Challenge would provide value to student pilots, provide another valuable tool in the instructor pilot's toolkit, and build a virtuous cycle of more and higher quality data enabling more and better training tools.

First, the ability to accurately identify maneuvers enables data infrastructure that can be used to further enhance training. As an example, every maneuver a student pilot flies in the simulator can be automatically identified and logged to allow for incredibly granular trend analysis across the pilot training enterprise. A student pilot would be able to quickly view every practice emergency landing pattern landing (ELP) he's flown over time and a flight commander overseeing 20 or more student pilots can view all of his students' ELPs in one place. Additionally, a dataset of all students' training which also contained labels of maneuvers would be significantly richer than the comparable raw data, as the data curation activities involved in this challenge demonstrate. Such a rich dataset spanning many years would contain enormous value and potential insights to everyone from instructor pilots to analysts at higher headquarters, to senior leaders.

Second, once maneuvers can be identified, the ability to grade maneuvers in an explainable way enables much earlier and more accessible learner-centered training, with more focused and effective use of instructor pilots' time resulting in more effective training. While athletes who become professional often begin practicing their sport in middle school or high school, Air Force pilots frequently receive no training or experience in flying until after college; this technology would help bridge that gap. With the low-cost of VR simulators, cadets in Reserve Officer Training Corps (ROTC) college programs, Air Force Academy cadets, and even Junior ROTC cadets in high school, who would normally have no access to flying training prior to arriving at Undergraduate Pilot Training after college, can begin learning basic flying fundamentals in VR simulators that will serve them well in pilot training. Student pilots who often wait several months or more for their formal pilot training start date can likewise access the same training.

Simulation software on such a ubiquitous VR simulator can demonstrate perfect maneuvers completed by instructors, grade early access students on maneuvers they attempt, and explain how to improve each maneuver to better emulate the instructor demonstration in an iterative and real-time way. The ability to identify and grade maneuvers can also be used in actual aircraft in real-time, as open mission systems capable of quickly deploying novel software on non-safety critical aircraft systems become more widely-available in military aircraft. This technology would accelerate debriefs and provide trend analysis and objective data back to even fully-qualified pilots to help improve their own flying. For example, a young fighter pilot wingman can compare and contrast how each of his maneuvers compares to those flown by the most experienced pilots in his squadron.

Finally, from a technology acquisition point of view, training-quality datasets and challenges enable companies hoping to provide such AI-enhanced pilot training technology to the government to objectively demonstrate their capabilities in comparison to benchmark algorithms and other industry and academia participants, making the market for such technology more open, objective, and competitive.

**ACKNOWLEDGEMENTS**

The authors wish to acknowledge the following individuals for their contributions and support: Sean Anderson, Ross Allen, Bob Bond, William Caballero, Jeff Gottschalk, Tucker Hamilton, Chris Hill, Nathan Gaw, Mike Kanaan, Tim Kraska, Charles Leiserson, Christian Prothmann, John Radovan, Steve Rejto, Daniela Rus, Allan Vanterpool, Marc Zissman, and the MIT SuperCloud team: Bill Arcand, Bill Bergeron, David Bestor, Chansup Byun, Nathan Frey, Michael Houle, Matthew Hubbell, Hayden Jananthan, Anna Klein, Joseph McDonald, Peter Michaleas, Julie Mullen, Andrew Prout, Antonio Rosa, Albert Reuther, Matthew Weiss. This research was sponsored by the United States Air Force Research Laboratory and the United States Air Force Artificial Intelligence Accelerator and was accomplished under Cooperative Agreement Number FA8750-19-2-1000. The views and conclusions contained in this document are those of the authors and should not be interpreted as representing the official policies, either expressed or implied, of the United States Air Force or the U.S. Government. The U.S. Government is authorized to reproduce and distribute reprints for Government purposes notwithstanding any copyright notation herein.